\documentclass{article} 
\usepackage{preprint,times}

\usepackage{url}
\usepackage{amsmath,amssymb,booktabs,microtype,multirow}
\usepackage{graphicx}
\usepackage{subcaption}
\usepackage[hidelinks]{hyperref}
\usepackage{flafter}
\usepackage{makecell}

\title{From Perception to Integration: \\ Revisiting the Internal Dynamics of Reasoning in Vision-Language Models}

\author{Rong Yu Xu$^{1}$, Prayag Tiwari$^{2}$ \& Shaolei Zhang$^{3}$\thanks{Corresponding author.} \\
{\small $^{1}$Shenzhen College of International Education \quad \texttt{s23372.xu@stu.scie.com.cn}} \\
{\small $^{2}$Halmstad University, Sweden \quad \texttt{prayag.tiwari@hh.se}} \\
{\small $^{3}$Renmin University of China \quad \texttt{zhangshaolei98@ruc.edu.cn}}
}

\begin{document}
\raggedbottom

\maketitle

\begin{abstract}
Vision-language models (VLMs) can answer simple visual questions, but often struggle when one question requires several visual judgments. We study this gap with controlled tasks for feature binding, numerosity, spatial relations, and amodal completion, together with a Composite task that combines them. Matched counterfactual image pairs isolate changes in the visual evidence needed to answer. Across four models, direct answers, hidden-state readouts, and state interventions show that the individual judgments can be made without explicit reasoning and that intervening on the corresponding states can affect the answer. During reasoning, the Composite answer becomes decodable from hidden states and usable from shortened traces, often before the model stops on its own. We train a small detector to predict this readiness and stop reasoning at that point. On MMStar and RealWorldQA, this reduces mean reasoning tokens by 79.1\% and 74.5\%, while average accuracy rises by 3.13 and 3.30 percentage points, respectively. These findings connect the internal development of answer readiness to a practical rule for allocating reasoning computation.
\end{abstract}

\begin{figure}[t]
    \centering
    \includegraphics[width=\linewidth]{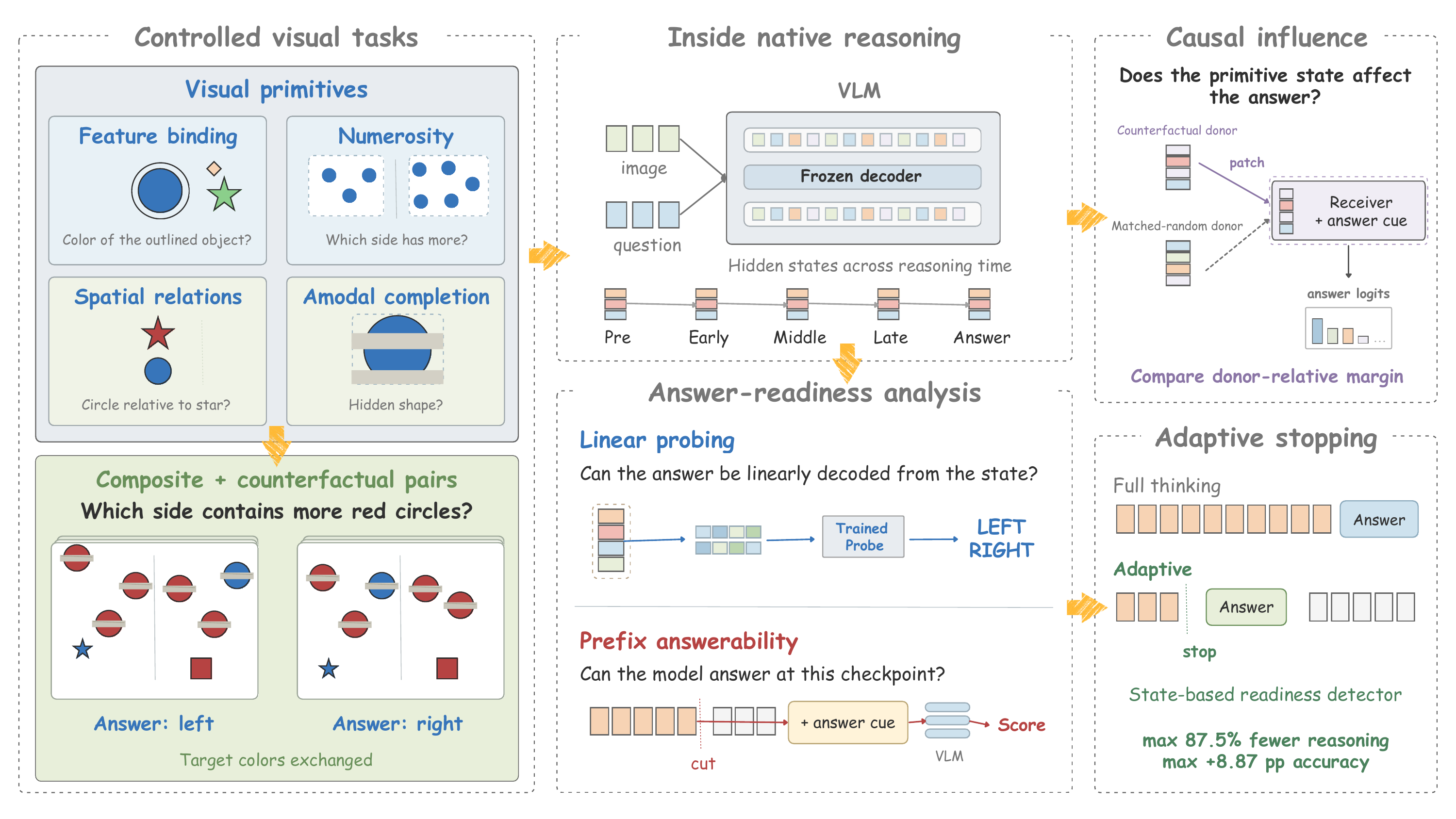}
    \caption{Overview of our study. Controlled visual tasks test individual judgments and their combination in Composite scenes. Linear readouts and counterfactual state interventions examine what visual information is represented and affects the answer. During native reasoning, readouts and answers from shortened traces track when the Composite answer becomes available. A hidden-state readiness detector uses this principle to stop reasoning early on MMStar and RealWorldQA.}
    \label{fig:overview}
\end{figure}

\section{Introduction}

People recognize many objects quickly, but some images require further visual processing. In primates, object information can appear within 150\,ms during an initial feedforward pass, while harder images require recurrent processing \citep{thorpe1996speed,dicarlo2012how,lamme2000distinct,kar2019evidence}. VLMs also have a fixed number of layers in each forward pass. Generating more tokens gives them additional steps of computation. This suggests a question: which visual judgments can a VLM make before it starts reasoning, and which improve as reasoning proceeds?

Visual benchmarks show which questions models answer correctly \citep{johnson2017clevr,huang2025visfactor}. Mechanistic studies show where visual information appears inside a model \citep{neo2025interpreting,zhang2025crossmodal}. We ask how the model uses that information to reach an answer (Figure~\ref{fig:overview}). Across four VLMs, accuracy without reasoning is 87.2--91.4\% on four separate visual tasks, compared with a 31.3\% guess rate. Yet average accuracy falls to 58.6\% on a single Composite question with two answer choices, left or right. This question requires the model to combine several visual judgments to select the correct side. High accuracy on the individual tasks therefore does not ensure a correct answer when those judgments must be used together. A correct final answer does not reveal when the model first became able to give it.

\begin{figure}[t]
    \centering
    \includegraphics[width=0.92\textwidth]{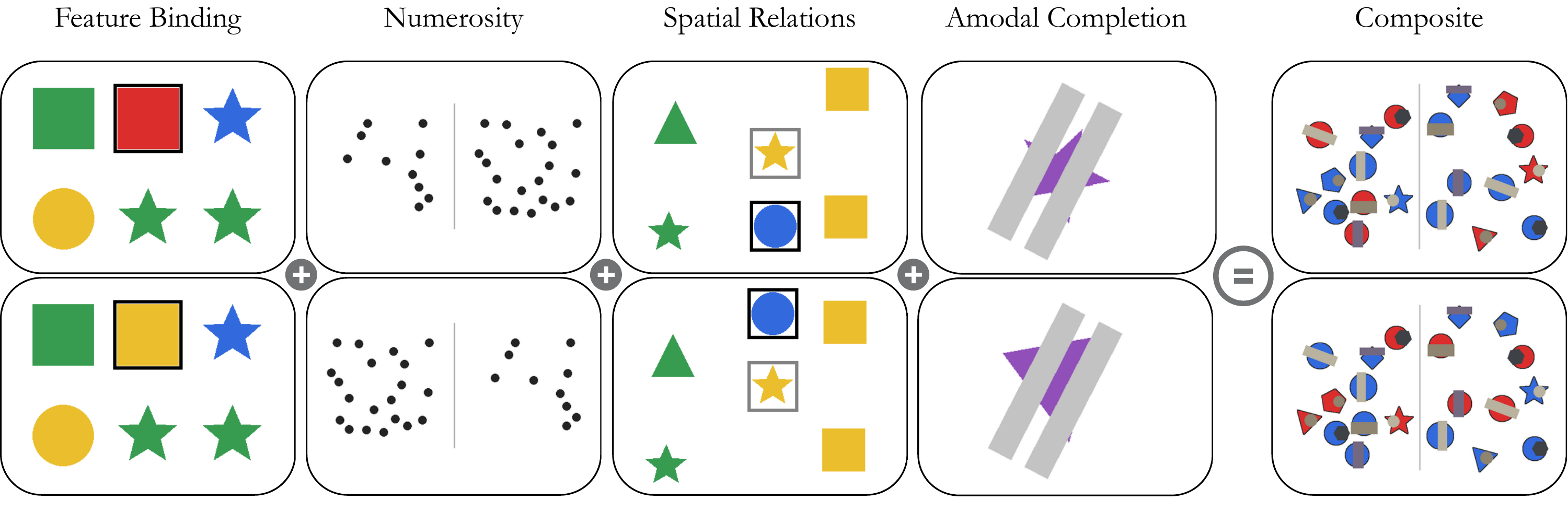}
    \caption{Every column shows a counterfactual pair in which the task-relevant variable and the correct answer change together while the remaining scene content is preserved.}
    \label{fig:dataset_examples}
\end{figure}

We test Feature Binding, Numerosity, Spatial Relations, and Amodal Completion: identifying an object's features, comparing quantities, judging positions, and recognizing a partly hidden shape \citep{treisman1980feature,feigenson2004core,logan1994spatial,kellman1991theory}. Our Composite task combines these demands by asking which side of a partly occluded scene contains more red circles (Figure~\ref{fig:dataset_examples}, rightmost column). We also use matched image pairs that change the correct answer while keeping the rest of the scene fixed \citep{gardner2020evaluating,bitton2021automatic}. Hidden-state readouts and state interventions test what information the model carries and uses. Answers from shortened reasoning traces show when it can use that information to answer the Composite question.

Reasoning raises Composite pair-consistent accuracy by 40.8 points on average. It is also costly: full traces average 1,650 reasoning tokens on MMStar, and 16.6\% of Composite runs reach the generation limit without a final answer. We find that a shortened trace can often support a correct answer before the model stops. We call the earliest tested point where this happens \emph{answer readiness} and ask whether the model's state can detect it during generation.

We propose Answer-Readiness-Guided Early Stopping. A linear detector uses the current hidden state and its change since the previous checkpoint to predict whether the model can answer correctly now. If the prediction crosses a threshold selected on validation data, the model stops reasoning and generates its answer. This approach builds on earlier methods that stop when an answer appears correct or sufficient \citep{zhang2025reasoningmodels,yang2026deer,xiang2026thinking}.

Across four VLMs on MMStar and RealWorldQA, stopping shortens mean reasoning by 79.1\% and 74.5\% while average accuracy rises by 3.13 and 3.30 points.

\section{Related Work}

\paragraph{Multimodal reasoning and visual evaluation.}
Controlled benchmarks test how VLMs combine visual information \citep{johnson2017clevr,schiappa2024probing,huang2025visfactor}. Multimodal chain-of-thought methods ask models to explain their answers or reason through image regions and objects \citep{zhang2024multimodalcot,shao2024visualcot,li2025vocot}. Other evaluations find that more reasoning helps some visual questions but not others \citep{jiang2025mmecot,jin2026looklight}. These studies measure final answers; they do not show when reasoning first makes an answer possible.

\paragraph{Interpreting visual representations in VLMs.}
Researchers use causal tracing and token interventions to locate information inside language models and VLMs \citep{meng2022locating,basu2024understanding,neo2025interpreting}. Related work studies how visual information moves through direct and text-mediated pathways \citep{zhang2025crossmodal,salazar2026pathways}. Layer-wise readouts also separate visual grounding from later processing and answer formatting \citep{yu2025multimodal}. These studies focus on a single forward pass. They do not track how the model uses visual information while generating a reasoning trace.

\paragraph{Efficient reasoning and early exit.}
Hidden-state readouts can predict reasoning success or check intermediate answers \citep{afzal2025knowing,zhang2025reasoningmodels}. Other methods stop when a trial answer has high confidence or appears sufficient \citep{yang2026deer,xiang2026thinking}. For VLMs, \citet{bi2026seeing} estimate when reasoning helps and how long it should continue, then intervene on attention heads. These methods do not use a direct measurement of when the visual answer becomes available during reasoning.

Our analysis measures when a Composite answer becomes usable during reasoning. That measurement provides the target for a detector that decides when to stop.

\section{Interpreting Composite Visual Reasoning}
\label{sec:interpretability}

\subsection{Interpretability Analysis Setup}
\label{sec:primitive}

Each task isolates one answer-determining variable while holding the remaining scene content constant.

\paragraph{Feature Binding.}
Feature Binding asks which color belongs to a marked object. People can notice a color and a shape but sometimes combine them incorrectly \citep{treisman1980feature,treisman1982illusory}. Each image contains several colored shapes; a black outline marks the target. The model reports its color.

\paragraph{Numerosity.}
Numerosity asks which side of a divider has more dots. Even infants can compare such quantities before learning number words \citep{feigenson2004core,xu2000large}. We vary the relationship between dot count and total dot area, so area alone does not always reveal the answer. The model responds left or right rather than giving an exact count.

\paragraph{Spatial Relations.}
Spatial Relations asks where a target lies relative to another object. Such judgments require selecting the two relevant objects \citep{carlson1999what,logan1994spatial}. Black and gray outlines mark the target and reference, while their colors and shapes vary. The model chooses left, right, above, or below.

\paragraph{Amodal Completion.}
Amodal Completion asks what whole shape is partly hidden behind an occluder. People can infer the whole from visible fragments \citep{kellman1991theory,sekuler1992perception}. We partly cover a geometric shape and ask the model to choose its complete form. The Composite task needs this judgment for partly hidden objects.

\paragraph{Composite Task.}
The Composite task asks which side of a divider has more red circles among colored, partly hidden shapes. To answer, the model must recognize each shape, match colors to shapes, assign objects to a side, and compare the two counts. This resembles ``challenge images'' in vision research: the parts may be easy to recognize, but combining them can require more processing \citep{kar2019evidence,kar2021fast}.

\paragraph{Counterfactual Pairs.}
For state interventions, we use matched image pairs that differ in the feature that determines the answer while keeping the rest of the scene fixed \citep{gardner2020evaluating,bitton2021automatic}. Feature Binding, Numerosity, and Spatial Relations use separate datasets for direct answers and interventions. Amodal Completion and Composite use paired datasets throughout. Figure~\ref{fig:dataset_examples} shows one pair per task; Appendix~\ref{app:data_generation} gives the construction rules and splits.

\subsection{Visual Judgments Without Reasoning}
\label{sec:answer_ready_states}

The four component judgments provide three complementary tests without explicit reasoning: direct-answer accuracy, linear readouts of hidden states, and interventions that measure whether changing those states affects the answer \citep{belinkov2021probing}.

\paragraph{Component judgments are available without reasoning.}
We freeze each model, disable thinking, and ask it to answer the four tasks directly. All four models perform above the task-specific guess levels (Table~\ref{tab:primitive_behavior}). They do best on Feature Binding and Spatial Relations and make more errors on Numerosity and Amodal Completion.

\begin{table}[!htbp]
    \centering
    \small
    \footnotesize
    \setlength{\tabcolsep}{4.5pt}
    \caption{All four component judgments can be performed above the guess level without explicit reasoning. Answer accuracy (\%) on 1,000 images per task and model; guess is uniform random selection over the answer choices (two for Numerosity, four for the others).}
    \label{tab:primitive_behavior}
    \begin{tabular}{lccccc}
        \toprule
        Task
        & Qwen3.5-4B
        & Qwen3.5-9B
        & Gemma-4-E4B
        & Gemma-4-12B
        & Guess \\
        \midrule
        Feature Binding
        & 98.0
        & 99.8
        & 98.6
        & 99.9
        & 25.0 \\
        Numerosity
        & 79.6
        & 85.2
        & 72.4
        & 86.5
        & 50.0 \\
        Spatial Relations
        & 99.8
        & 100.0
        & 97.5
        & 89.5
        & 25.0 \\
        Amodal Completion
        & 76.8
        & 80.5
        & 81.4
        & 72.9
        & 25.0 \\
        \midrule
        \textbf{Avg}
        & \textbf{88.6}
        & \textbf{91.4}
        & \textbf{87.5}
        & \textbf{87.2}
        & 31.3 \\
        \bottomrule
    \end{tabular}
\end{table}

\begin{figure}[t]
    \centering
    \includegraphics[width=\linewidth]{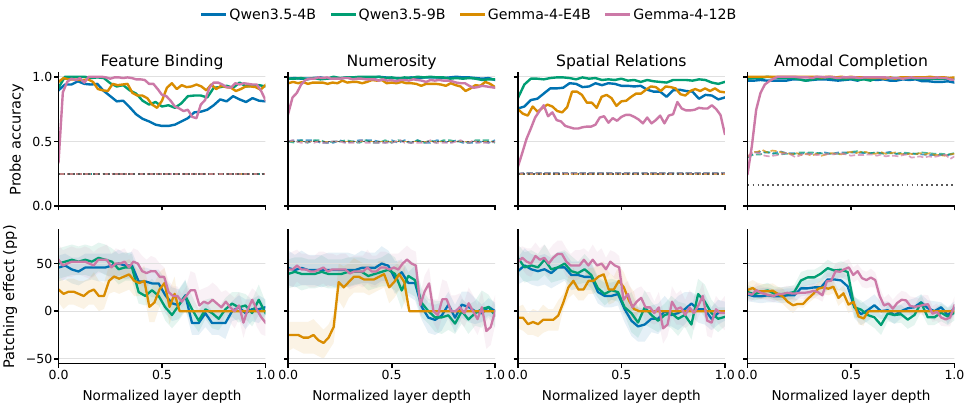}
    \caption{Top: test accuracy of readouts using image states (solid) or question-only states (dashed); gray dotted lines mark chance. Bottom: how often matched donor states move scores toward the donor's answer compared with control donors, in percentage points. Shading shows pair-bootstrap 95\% CIs. Layer depth is normalized within each model.}
    \label{fig:primitive_analysis}
\end{figure}

\paragraph{Primitive variables are linearly recoverable across depth.}
Direct answers show that a model can make these judgments, but do not show where it represents them. We therefore average its image-token hidden states at each layer:
\begin{equation}
    v_i^{(\ell)}=\frac{1}{|\mathcal I_i|}
    \sum_{t\in\mathcal I_i}h_{i,t}^{(\ell)},
\end{equation}
where $h_{i,t}^{(\ell)}$ is the representation of sample $i$ at layer $\ell$ and token position $t$, and $\mathcal I_i$ contains its image-token positions. We standardize these vectors using training-set statistics to obtain $z_i^{(\ell)}$, whose rows form the training matrix $Z_\ell$. For each model and task, we fit an independent ridge linear readout at every layer, keeping the VLM frozen and fitting only the readout matrix:
\begin{equation}
    W_\ell^*=\arg\min_W\|Z_\ell W-Y\|_F^2+\lambda\|W\|_F^2,
    \qquad
    f_{\mathrm{read}}\bigl(z_i^{(\ell)}\bigr)=\arg\max_c\bigl(z_i^{(\ell)}W_\ell^*\bigr)_c.
\end{equation}
Here $Y$ contains the correct task label: color, side with more dots, spatial direction, or complete shape. The parameter $\lambda$ controls regularization, and $f_{\mathrm{read}}$ is the fitted readout. The accuracy of $f_{\mathrm{read}}$ measures how well that label can be decoded from the frozen model's states.

We test each readout on held-out samples and keep paired images in the same split. As a control, we fit a separate readout using question tokens without an image. For Amodal Completion, the readout predicts one of six shape labels, whereas direct answering chooses among four options in the question. Appendix~\ref{app:primitive_analysis} gives the splits and fitting details.

In the upper row of Figure~\ref{fig:primitive_analysis}, each task label is decodable from several layers. Numerosity and complete-shape accuracy is high; Feature Binding and Spatial Relations vary more across layers. Question-only controls stay near chance for the first three tasks. They score higher for Amodal Completion because the candidate list itself gives some information about the shape. Image-conditioned readouts outperform these controls through most layers. State interventions test whether this decoded information also affects the answer.

\paragraph{State interventions affect answers at some layers.}
Reading a task label from a state does not prove that the model uses that state to answer. We test this with matched image pairs. One image is the receiver; the other, with a different correct answer, is the donor. At a selected decoder block, we replace the receiver's image-token states with the donor's and continue the forward pass:
\begin{equation}
    f_{\mathrm{patch}}(r,d,\ell)\quad\text{assigns}\quad
    H_{r,\mathcal I_r}^{(\ell)}\leftarrow H_{d,\mathcal I_d}^{(\ell)}.
\end{equation}
We swap states in both directions and retain only pairs for which both original answers are correct under the intervention scoring protocol. This leaves a small set whose size varies by model. As a control, we also use a donor from another pair that shares the receiver's label.

An intervention can shift answer scores even if the top answer stays the same \citep{heimersheim2024activation}. We measure this shift with $m=s(y_d)-s(y_r)$, the score for the donor's answer minus the score for the receiver's answer. We then compare how often the matched donor and the control donor increase this margin:
\begin{equation}
    \Delta_{\mathrm{patch}}=100\left[
    \Pr(m_{\mathrm{target}}>m_{\mathrm{base}})
    -\Pr(m_{\mathrm{random}}>m_{\mathrm{base}})
    \right].
\end{equation}
Here $m_{\mathrm{base}}$ is the receiver's score before replacement. A positive $\Delta_{\mathrm{patch}}$ means that matched donors push scores toward their answers more often than control donors do. It measures how often scores move, not how far they move or how often the final choice flips. Appendix~\ref{app:primitive_analysis} gives the scoring and matching rules.

In the lower row of Figure~\ref{fig:primitive_analysis}, matched donors move scores toward their answers more often than control donors in early and middle layers, but less consistently at the final tested block. Some task labels remain decodable in late layers even when matched-donor replacement no longer shows a consistent advantage over control replacement. Thus, decodability and the measured intervention effect can differ across depth. The results suggest that the Composite task is difficult because the model must combine the component judgments, not simply recognize their parts. A readout alone cannot establish that the model will use the information. We then test whether a shortened reasoning trace already supports the correct answer.

\subsection{Development of Composite Answer Readiness}

\paragraph{Reasoning improves counterfactual consistency.}
We compare direct answers with the model's normal reasoning on the same held-out Composite pairs. Sample accuracy counts correct images. Pair-consistent accuracy counts a pair only when both images are answered correctly.

\begin{table}[!htbp]
    \centering
    \small
    \renewcommand{\arraystretch}{1.12}
    \setlength{\tabcolsep}{2pt}
    \caption{Native thinking improves counterfactual consistency. Values are accuracy (\%) with pair-bootstrap 95\% CIs; $\Delta$ is the reasoning-induced change in percentage points. Budget hit counts thinking-mode runs that reach the generation limit, out of 200 per model; the summary row pools all 800 runs.}
    \label{tab:thinking_behavior}
    \begin{tabular*}{\textwidth}{
        @{\extracolsep{\fill}}
        l
        ccc
        ccc
        c@{}
    }
        \toprule
        & \multicolumn{3}{c}{\textbf{Sample Accuracy}}
        & \multicolumn{3}{c}{\textbf{Pair-Consistent Accuracy}} & \textbf{Budget hit} \\
        \cmidrule(lr){2-4}
        \cmidrule(lr){5-7}\cmidrule(lr){8-8}
        Model
        & Non-thinking
        & \textbf{Thinking}
        & $\Delta$
        & Non-thinking
        & \textbf{Thinking}
        & $\Delta$ & Count (\%) \\
        \midrule

        Qwen3.5-4B
        & \makecell{58.5\\{\footnotesize [53.0,64.0]}}
        & 
          \makecell{\textbf{80.0}\\{\footnotesize [74.0,86.0]}}
        & $\uparrow\,21.5$
        & \makecell{26.0\\{\footnotesize [18.0,35.0]}}
        & 
          \makecell{\textbf{67.0}\\{\footnotesize [58.0,76.0]}}
        & $\uparrow\,41.0$ & 38 (19.0\%) \\

        Qwen3.5-9B
        & \makecell{66.5\\{\footnotesize [61.5,71.5]}}
        & 
          \makecell{\textbf{94.0}\\{\footnotesize [90.5,97.0]}}
        & $\uparrow\,27.5$
        & \makecell{36.0\\{\footnotesize [27.0,45.0]}}
        & 
          \makecell{\textbf{89.0}\\{\footnotesize [83.0,95.0]}}
        & $\uparrow\,53.0$ & 11 (5.5\%) \\

        Gemma-4-E4B
        & \makecell{52.0\\{\footnotesize [49.0,55.0]}}
        & 
          \makecell{\textbf{73.5}\\{\footnotesize [67.0,80.0]}}
        & $\uparrow\,21.5$
        & \makecell{7.0\\{\footnotesize [2.0,12.0]}}
        & 
          \makecell{\textbf{57.0}\\{\footnotesize [47.0,67.0]}}
        & $\uparrow\,50.0$ & 0 (0.0\%) \\

        Gemma-4-12B
        & \makecell{\textbf{57.5}\\{\footnotesize [54.0,61.0]}}
        & 
          \makecell{54.5\\{\footnotesize [47.0,62.0]}}
        & $\downarrow\,3.0$
        & \makecell{15.0\\{\footnotesize [8.0,22.0]}}
        & 
          \makecell{\textbf{34.0}\\{\footnotesize [25.0,43.0]}}
        & $\uparrow\,19.0$ & 84 (42.0\%) \\

        \midrule
        \textbf{Avg}
        & 58.6
        & \textbf{75.5}
        & $\uparrow\,16.9$
        & 21.0
        & \textbf{61.8}
        & $\uparrow\,40.8$ & 133 (16.6\%) \\

        \bottomrule
    \end{tabular*}
\end{table}

Reasoning improves pair-consistent accuracy for all four models (Table~\ref{tab:thinking_behavior}). Sample accuracy rises for three models but falls for Gemma-4-12B. This model also often reaches the generation limit; these results do not show whether that limit causes its lower sample accuracy. Final answers still leave one question open: when during reasoning could the model first give the correct answer?

\paragraph{Tracing when the Composite answer becomes usable.}
We examine each model's reasoning trace at five checkpoints: before reasoning (\textsc{Pre}), early, middle, at the last reasoning token (\textsc{Late}), and just before the final choice (\textsc{Answer}). Validation data select one layer per model based on how well it represents the answer before reasoning. We use that same layer at every checkpoint and on all test samples. This analysis includes only pairs with valid checkpoints and parsed answers in both traces, so its sample sizes can differ from the direct-answer test. Appendix~\ref{app:reasoning_checkpoints} defines the checkpoints and selected layers; Appendix~\ref{app:data_generation} gives the splits.

At each checkpoint, a separate linear readout predicts the correct side from the selected hidden state. We use the same feature standardization, ridge objective, and pair-preserving splits as in Section~\ref{sec:answer_ready_states}. Test accuracy measures how well the answer is decodable from that state.

We also test whether the model can answer from the reasoning generated so far. At each checkpoint, we truncate the trace after that checkpoint, append the same answer cue, and score the two possible answers with the frozen model:
\begin{equation}
    f_{\mathrm{force}}(i,t)
    =\arg\max_{c\in\{\mathrm{left},\mathrm{right}\}}
    s_\theta(c\mid x_i,r_{i,\leq t},a),
    \qquad
    U(t)=\frac{1}{N}\sum_{i=1}^{N}
    \mathbf{1}\bigl[f_{\mathrm{force}}(i,t)=y_i\bigr].
\end{equation}
Here $x_i$ is the image and prompt, $r_{i,\leq t}$ is the reasoning kept through checkpoint $t$, and $a$ is the fixed answer cue, which gives the format but not the answer. The score $s_\theta$ compares the two choices. Unlike the readout from one hidden state, $f_{\mathrm{force}}$ uses the full retained trace, including any intermediate reasoning in text. Its accuracy measures whether the answer is usable at each checkpoint. This offline test needs the correct answer for scoring; that answer is unavailable when the model is deployed.

\begin{figure}[!htbp]
    \centering
    \includegraphics[width=\linewidth]{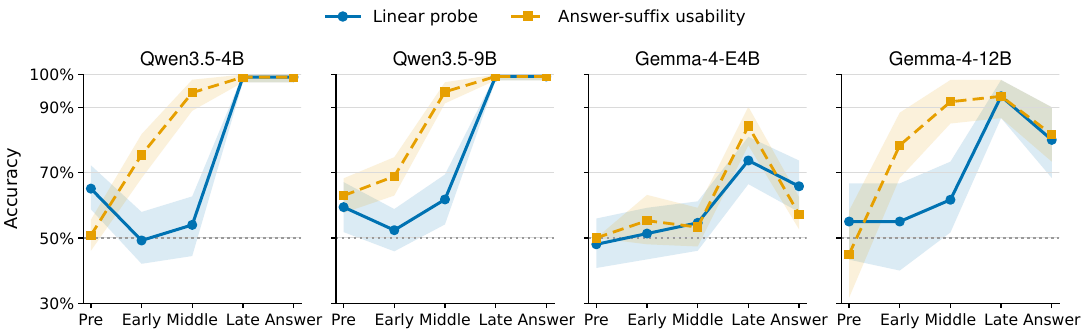}
    \caption{Each panel shows one model at five reasoning checkpoints. Blue: accuracy of a hidden-state readout for the Composite answer. Orange: accuracy when the model answers from a trace cut at that checkpoint. Shading shows pair-bootstrap 95\% CIs; the dotted line marks chance.}
    \label{fig:answer_state_dynamics}
\end{figure}

\paragraph{Decodability and usability develop during reasoning.}
Before reasoning, neither the hidden-state readout nor the shortened-trace test is consistently accurate across models (Figure~\ref{fig:answer_state_dynamics}). Both improve during reasoning, though at different speeds. Usability rises earlier for the Qwen models and Gemma-4-12B, and later for Gemma-4-E4B. By \textsc{Late}, both decodability and usability are high for all four models.

Accuracy does not always improve from \textsc{Late} to \textsc{Answer}: it stays high for the Qwen models but falls for both Gemma models. Correct answers are often available at earlier checkpoints, before the model finishes reasoning. Moreover, 16.6\% of Composite thinking runs reach the generation limit without a final answer (Table~\ref{tab:thinking_behavior}). These results motivate predicting usability during generation.

\section{Answer-Readiness-Guided Early Stopping}
\label{sec:early_stopping}

The shortened-trace test shows that an answer can be usable before reasoning ends. Figure~\ref{fig:overview} shows how this finding becomes a stopping rule. For each model and dataset, a small detector reads the current hidden state and its change at a layer selected on Composite data. Its predicted readiness score determines when to stop: once it crosses a threshold chosen on validation data, the model generates its answer. This resembles the use of an evidence threshold in studies of human perceptual decisions \citep{oconnell2012supramodal}.

\paragraph{Predicting readiness without the correct answer.}
The offline test $f_{\mathrm{force}}$ uses the correct answer to measure readiness. At inference time, the detector must predict readiness without that answer. For training, we stop recorded traces at fixed checkpoints, ask the frozen VLM to answer, and label each checkpoint by whether that answer is correct. The detector uses the current state, its change since the previous checkpoint, and the elapsed reasoning length:
\begin{equation}
    f_{\mathrm{state}}(i,t)=\bigl[h_{i,t}^{(\ell)},\ d_{i,t}^{(\ell)},\ t/B\bigr],
    \qquad
    f_{\mathrm{ready}}(i,t)=\sigma\bigl(w^\top\tilde f_{\mathrm{state}}(i,t)+b\bigr),
\end{equation}
Here $t$ counts generated reasoning tokens, $d_{i,t}^{(\ell)}$ is the change since the previous checkpoint, and $B$ is the generation budget. At the first checkpoint, we use a zero vector as the previous state, so $d_{i,\Delta}^{(\ell)}=h_{i,\Delta}^{(\ell)}$; subsequent checkpoints use $d_{i,t}^{(\ell)}=h_{i,t}^{(\ell)}-h_{i,t-\Delta}^{(\ell)}$. Training and online inference use the same convention. The tilde denotes standardization using training data. Only the detector weights $w$ and bias $b$ are trained.

\paragraph{Detector training.}
With checkpoint labels $y_{i,t}=\mathbf{1}[\hat a_{i,t}^{\mathrm{forced}}=a_i^*]$, we minimize
\begin{equation}
    \mathcal L=-\frac{1}{M}\sum_{(i,t)\in\mathcal D_{\mathrm{train}}}
    \bigl[\alpha y_{i,t}\log f_{\mathrm{ready}}(i,t)
    +(1-y_{i,t})\log\bigl(1-f_{\mathrm{ready}}(i,t)\bigr)\bigr],
\end{equation}
where $\hat a_{i,t}^{\mathrm{forced}}$ is the forced answer, $a_i^*$ is its reference, $M$ counts training checkpoints, and $\alpha$ is their negative-to-positive label ratio. Feature normalization and class weights use only training checkpoints.

\paragraph{Stopping at a validated decision bound.}
We check the detector every $\Delta=256$ reasoning tokens. At the first checkpoint where $f_{\mathrm{ready}}(i,t)\geq\eta$, we close the reasoning channel and ask the VLM for its final answer. The interval is fixed; validation data select only the threshold $\eta$. We choose the threshold that minimizes mean reasoning length while keeping accuracy within a set tolerance of full thinking. If the score never crosses it, generation continues until the model stops naturally or reaches the token budget.

\section{Experiments}
\label{sec:experiments}

\subsection{Experimental Setup}

We compare each model's adaptive stopping run with its full-thinking run using the same token budget and answer extraction rule. MMStar \citep{chen2024we} has multiple-choice questions about perception, science, and mathematics. RealWorldQA \citep{xai2024realworldqa} uses photographs and includes both multiple-choice and short-answer questions. We test all 1,500 MMStar and 765 RealWorldQA questions with two-fold cross-validation. One fold supplies detector training and threshold validation data; the other is held out for testing. All checkpoints from one question, and RealWorldQA questions with the same image, stay in the same fold. Each question therefore receives one held-out prediction per condition. Both conditions use a $B=6{,}144$-token budget. Appendix~\ref{app:early_stopping} gives the split counts.

\paragraph{Threshold selection and answer transition.}
Validation selects the threshold that uses the fewest mean reasoning tokens while allowing at most 0.5 percentage points less accuracy than full thinking. When the detector triggers, we close the thinking channel and append \texttt{Therefore, the answer is}. Training labels and adaptive inference use this same transition. Each model, dataset, and evaluation direction has its own detector and threshold.

\paragraph{Answer scoring and token accounting.}
Within each benchmark, both conditions use the same answer extraction rule, which does not depend on the correct answer. MMStar uses the final choice letter. RealWorldQA accepts a choice or short answer, as the question requires. We extract answers after the thinking channel closes and also accept direct short answers for RealWorldQA. A missing or unparseable answer counts as wrong.

Reasoning length counts tokens before the thinking channel closes. If it never closes, we count all generated tokens as reasoning. A direct answer has zero reasoning tokens. Total output length also includes answer tokens. We average lengths over all held-out questions, including those answered before the first checkpoint.

\subsection{Main Results}

\paragraph{Shorter reasoning with higher average accuracy.}
On MMStar, every model uses fewer reasoning tokens with adaptive stopping, and average accuracy rises (Table~\ref{tab:early_stopping_main}). Gemma-4-12B gains the most accuracy (+8.87 points), while Qwen3.5-4B is nearly unchanged (+0.07). Total output tokens fall by 23.3--84.9\% across models (Table~\ref{tab:output_stopping_stats}). The detector stops at the first 256-token check on 15.9--94.7\% of questions, depending on the model. Appendix~\ref{app:early_stopping_realworldqa} reports each test fold separately.

\begin{table}[!htbp]
    \centering
    \footnotesize
    \renewcommand{\arraystretch}{1.08}
    \setlength{\tabcolsep}{2pt}
    \caption{Adaptive stopping cuts reasoning cost on both benchmarks while raising average accuracy. All held-out cases per model (1,500 MMStar and 765 RealWorldQA); accuracy in \%, reasoning tokens as means, $\Delta$ in percentage points, reduction as the token decrease.}
    \label{tab:early_stopping_main}
    \begin{tabular*}{\textwidth}{
        @{\extracolsep{\fill}}
        ll
        ccc
        ccc@{}
    }
        \toprule
        & &  \multicolumn{3}{c}{\textbf{Accuracy (\%)}}
        & \multicolumn{3}{c}{\textbf{Reasoning Tokens}} \\
        \cmidrule(lr){3-5}
        \cmidrule(lr){6-8}
        Benchmark
        & Model
        & Full
        & \textbf{Stop}
        & $\Delta$
        & Full
        & \textbf{Stop}
        & Reduction \\
        \midrule
        \multirow{5}{*}{MMStar}
        & Qwen3.5-4B & 64.87 & \textbf{64.93} & $\uparrow\,0.07$ & 2,041 & \textbf{255} & $87.5\%$ \\

        & Qwen3.5-9B & 66.87 & \textbf{69.87} & $\uparrow\,3.00$ & 1,951 & \textbf{255} & $86.9\%$ \\

        & Gemma-4-E4B & 58.07 & \textbf{58.67} & $\uparrow\,0.60$ & 753 & \textbf{585} & $22.3\%$ \\

        & Gemma-4-12B & 56.93 & \textbf{65.80} & $\uparrow\,8.87$ & 1,854 & \textbf{286} & $84.6\%$ \\
        \cmidrule(l){2-8}
        & \textbf{Avg} & 61.69 & \textbf{64.82} & $\uparrow\,3.13$ & 1,650 & \textbf{345} & $\mathbf{79.1\%}$ \\
        \midrule
        \multirow{5}{*}{RealWorldQA}
        & Qwen3.5-4B & 72.81 & \textbf{75.42} & $\uparrow\,2.61$ & 1,696 & \textbf{285} & $83.2\%$ \\
        
        & Qwen3.5-9B & 72.94 & \textbf{77.91} & $\uparrow\,4.97$ & 1,606 & \textbf{248} & $84.6\%$ \\
        
        & Gemma-4-E4B & \textbf{58.17} & 57.12 & $\downarrow\,1.05$ & 442 & \textbf{346} & $21.8\%$ \\
        
        & Gemma-4-12B & 60.13 & \textbf{66.80} & $\uparrow\,6.67$ & 859 & \textbf{293} & $65.9\%$ \\
        \cmidrule(l){2-8}
        & \textbf{Avg} & 66.01 & \textbf{69.31} & $\uparrow\,3.30$ & 1,151 & \textbf{293} & $\mathbf{74.5\%}$ \\
        \bottomrule
    \end{tabular*}
\end{table}

\pagebreak
\paragraph{Results on short-answer questions.}
RealWorldQA tests whether the rule also works with short answers (Table~\ref{tab:early_stopping_main}). All four models use fewer reasoning tokens, and total output falls by 19.9--83.7\% (Table~\ref{tab:output_stopping_stats}). Average accuracy rises, although Gemma-4-E4B loses 1.05 points. The rule reduces output length for both answer formats; its accuracy effect varies by model.

\begin{table}[!htbp]
\centering\footnotesize
\renewcommand{\arraystretch}{1.08}
\setlength{\tabcolsep}{4pt}
    \caption{Mean total output tokens and detector-triggered stopping (\% of all questions).}
\label{tab:output_stopping_stats}
\begin{tabular*}{\textwidth}{@{\extracolsep{\fill}}lrrrrr@{}}
\toprule
& \multicolumn{3}{c}{Total output tokens} & \multicolumn{2}{c}{Stopping (\%)} \\
\cmidrule(lr){2-4}\cmidrule(l){5-6}
Model & Full & Stop & Reduction & Any check & First check \\
\midrule
\multicolumn{6}{l}{\textit{MMStar}} \\
Qwen3.5-4B & 2,159 & 326 & 84.9\% & 95.1\% & 94.7\% \\
Qwen3.5-9B & 2,066 & 315 & 84.8\% & 94.7\% & 94.3\% \\
Gemma-4-E4B & 944 & 724 & 23.3\% & 24.7\% & 15.9\% \\
Gemma-4-12B & 1,974 & 351 & 82.2\% & 76.3\% & 68.1\% \\
\midrule
\multicolumn{6}{l}{\textit{RealWorldQA}} \\
Qwen3.5-4B & 1,699 & 299 & 82.4\% & 79.1\% & 71.1\% \\
Qwen3.5-9B & 1,610 & 263 & 83.7\% & 79.3\% & 78.4\% \\
Gemma-4-E4B & 448 & 359 & 19.9\% & 38.2\% & 29.5\% \\
Gemma-4-12B & 865 & 314 & 63.7\% & 39.9\% & 33.1\% \\
\bottomrule
\end{tabular*}
\end{table}

\paragraph{Where the accuracy gains come from.}
Table~\ref{tab:completion_counts} separates questions by whether full thinking closes its reasoning channel. On MMStar, adaptive stopping reduces missing answers from 297 to 24 for Qwen3.5-4B and from 276 to 44 for Qwen3.5-9B. For Qwen3.5-4B the 96 correct answers gained on open traces are almost exactly offset by 95 lost on closed traces, leaving a net gain of one; Qwen3.5-9B gains 121 on open traces against 76 lost on closed ones. Gemma-4-12B gains on both groups: 100 correct answers on open traces and 33 on closed ones, while Gemma-4-E4B gains 20 on open traces and loses 11 on closed ones. On RealWorldQA, Gemma-4-E4B has no open full-thinking traces to recover; it loses eight correct answers on closed traces.

\begin{table}[!htbp]
\centering\small
\renewcommand{\arraystretch}{1.12}
\setlength{\tabcolsep}{3pt}
\caption{Answer completion and changes in correct-answer counts. Missing includes traces without a parsed answer. Open/closed refer to full thinking; direct answers count as closed. Gains are stopping minus full thinking.}
\label{tab:completion_counts}
\begin{tabular*}{\textwidth}{@{\extracolsep{\fill}}lrrrrrrr@{}}
\toprule
& \multicolumn{2}{c}{Missing answer} & \multicolumn{2}{c}{Full status} & \multicolumn{3}{c}{Net correct gain} \\
\cmidrule(lr){2-3}\cmidrule(lr){4-5}\cmidrule(l){6-8}
Model & Full & Stop & Unclosed & Closed & Unclosed & Closed & Total \\
\midrule
\multicolumn{8}{l}{\textit{MMStar}} \\
Qwen3.5-4B & 297 & 24 & 267 & 1,233 & 96 & $-95$ & 1 \\
Qwen3.5-9B & 276 & 44 & 243 & 1,257 & 121 & $-76$ & 45 \\
Gemma-4-E4B & 104 & 87 & 32 & 1,468 & 20 & $-11$ & 9 \\
Gemma-4-12B & 318 & 56 & 207 & 1,293 & 100 & 33 & 133 \\
\midrule
\multicolumn{8}{l}{\textit{RealWorldQA}} \\
Qwen3.5-4B & 112 & 8 & 112 & 653 & 48 & $-28$ & 20 \\
Qwen3.5-9B & 96 & 12 & 96 & 669 & 53 & $-15$ & 38 \\
Gemma-4-E4B & 3 & 29 & 0 & 765 & 0 & $-8$ & $-8$ \\
Gemma-4-12B & 25 & 20 & 16 & 749 & 9 & 42 & 51 \\
\bottomrule
\end{tabular*}
\end{table}

\section{Conclusion}\label{sec:conclusion}

VLMs can make the component visual judgments without explicit reasoning, but combining them is harder. During reasoning, the Composite answer becomes decodable from hidden states and usable from a shortened trace, often before the model stops on its own. A small detector predicts this point and stops reasoning early. Across MMStar and RealWorldQA, it reduces reasoning tokens and improves average accuracy, partly by producing answers from traces that otherwise end without one.

\bibliography{main}
\bibliographystyle{plainnat}

\appendix
\section{Datasets and Mechanistic Analysis}
\label{app:data_generation}

Table~\ref{tab:dataset_splits} summarizes the synthetic datasets and their analysis splits.

\begin{table}[!htbp]
    \centering
    \small
    \renewcommand{\arraystretch}{1.05}
    \setlength{\tabcolsep}{2pt}
    \caption{\textbf{Synthetic dataset sizes and analysis splits.} Counts are images. Primitive splits are used for linear probing, and the Composite split is used for reasoning-trajectory analysis before trace filtering. Behavioral evaluation uses all 1,000 images per primitive and the 200-image Composite test set. Paired images remain in the same split.}
    \label{tab:dataset_splits}
    \begin{tabular*}{\textwidth}{@{\extracolsep{\fill}}lcccccc@{}}
        \toprule
        & & & \multicolumn{3}{c}{\textbf{Analysis Split}} & \\
        \cmidrule(lr){4-6}
        Dataset & Images & Choices & Train & Validation & Test & Split Unit \\
        \midrule
        Feature Binding & 1,000 & 4 & 800 & -- & 200 & Image \\
        Numerosity & 1,000 & 2 & 800 & -- & 200 & Image \\
        Spatial Relations & 1,000 & 4 & 800 & -- & 200 & Image \\
        Amodal Completion & 1,000 & 4 & 800 & -- & 200 & Pair \\
        Composite & 1,000 & 2 & 640 & 160 & 200 & Pair \\
        \bottomrule
    \end{tabular*}
\end{table}

\subsection{Primitive Dataset Construction}

We generate each dataset so the feature that determines the answer varies while other scene details are controlled. The settings for each task follow.

\paragraph{Feature Binding.}
Feature Binding uses four target colors (red, blue, green, and yellow), four shapes, 4--8 objects per image, and three layout templates. Target shape, size, and position also vary.

\paragraph{Numerosity.}
Numerosity uses three ratios between the larger and smaller dot counts (2.0, 1.5, and 1.25). For each ratio, total dot area either agrees with the count difference, is similar on both sides, or points in the opposite direction. Dots do not overlap, and each side is equally often the one with more dots.

\paragraph{Spatial Relations.}
Spatial Relations uses black and gray outlines for the target and reference, with their colors and shapes sampled independently. Its conditions cross four directions, three target--reference distances (60, 90, and 120 pixels), and three scene sizes (4, 6, and 8 objects).

\paragraph{Amodal Completion.}
Amodal Completion uses six shapes and six kinds of occluder. Each shape is 25--65\% hidden but has at least two visible parts. The two images in a pair have different complete shapes. They share the target's color, center, size, and rotation settings, the occluder, the intended amount of occlusion, and the order of four answer choices. The actual hidden area can differ between the shapes.

\paragraph{Behavioral and probe splits.}
The VLM stays frozen during direct answering and readout fitting. We balance splits by target color for Feature Binding; by count ratio, area condition, and answer for Numerosity; and by relation, distance, and object count for Spatial Relations. Amodal Completion has 500 matched pairs: 400 for training and 100 for testing, with occluder types balanced. Paired images always stay in the same split.

\paragraph{Counterfactual pairs for primitive interventions.}
Feature Binding, Numerosity, and Spatial Relations each have a separate intervention set of 125 matched pairs: 100 training and 25 test pairs. Each pair keeps the scene context fixed while changing the feature that determines the answer. Amodal interventions use the 100 held-out Amodal Completion pairs, which change the complete shape under the same occluder. We intervene in both directions and apply the criteria in Section~\ref{sec:answer_ready_states} to choose pairs for the primary analysis.

\subsection{Primitive Readouts and Interventions}
\label{app:primitive_analysis}

This subsection gives the settings for Section~\ref{sec:answer_ready_states}. Direct-answer tests disable thinking, request a short response, and allow at most 40 generated tokens. We accept a normalized choice letter or a matching word answer. Table~\ref{tab:primitive_readout_settings} summarizes the readouts. For standardized training features $X$ and one-hot labels $Y$, we solve
\[
W=X^\top(XX^\top+100I)^{-1}Y,
\]
and predict the label maximizing $zW$ for a standardized test vector $z$.

\begin{table}[!htbp]
\centering\small
\renewcommand{\arraystretch}{1.05}
\setlength{\tabcolsep}{4pt}
\caption{\textbf{Primitive linear-readout settings.} Dataset splits are given in Table~\ref{tab:dataset_splits}.}
\label{tab:primitive_readout_settings}
\begin{tabular*}{\textwidth}{@{\extracolsep{\fill}}p{0.25\textwidth}p{0.71\textwidth}@{}}
\toprule
Setting & Specification \\
\midrule
Inputs & No-thinking image and question \\
Visual features & Mean-pooled image-token states at each tested depth \\
Question-only control & Mean-pooled question span, including candidates, with no image \\
Standardization & Per-dimension training mean and standard deviation \\
Label classes & 4 for Feature Binding, 2 for Numerosity, 4 for Spatial Relations, and 6 for Amodal Completion \\
\bottomrule
\end{tabular*}
\end{table}

\paragraph{Patching protocol.}
The no-thinking prompt ends in an answer cue, and we score the next-token choice. Each choice score is the maximum logit over the last token IDs of uppercase/lowercase and leading-space variants of its letter. Pairs and controls satisfy the following criteria:
\begin{itemize}
\item Both intervention directions are available and both unpatched answers are correct.
\item The original donor and receiver margins differ by more than $10^{-6}$ in both directions.
\item Matched-random donors come from the same test set, belong to another pair, and share the receiver's semantic label. Amodal controls additionally match question-token length.
\end{itemize}
The lower row of Figure~\ref{fig:primitive_analysis} averages the two directional indicator differences within each pair. Table~\ref{tab:patch_retained_pairs} gives the retained sample sizes.

\begin{table}[!htbp]
\centering\small
\renewcommand{\arraystretch}{1.05}
\setlength{\tabcolsep}{4pt}
\caption{\textbf{Retained counterfactual pairs for primitive interventions.}}
\label{tab:patch_retained_pairs}
\begin{tabular*}{\textwidth}{@{\extracolsep{\fill}}lrrrr@{}}
\toprule
Task & Qwen3.5-4B & Qwen3.5-9B & Gemma-4-E4B & Gemma-4-12B \\
\midrule
Feature Binding & 24 & 25 & 22 & 25 \\
Numerosity & 22 & 23 & 18 & 24 \\
Spatial Relations & 25 & 25 & 22 & 25 \\
Amodal Completion & 48 & 62 & 72 & 75 \\
\bottomrule
\end{tabular*}
\end{table}

\subsection{Composite Data and Reasoning Checkpoints}
\label{app:reasoning_checkpoints}

The Composite analysis uses scenes that require several component judgments. We also define fixed points where we can cut off a reasoning trace and test whether the model can answer.

\paragraph{Composite dataset.}
The Composite task has 1,000 images in 500 matched pairs. Each side shows 8--12 partly hidden red or blue shapes; the model must choose the side with more red circles. Within a pair, we change which shapes have which colors to reverse the answer. Positions, shapes, sizes, rotations, occluders, and overall color and shape counts remain fixed. Occlusion ranges from 25\% to 60\%. We reserve 100 pairs (200 images) for direct-answer testing. The other 400 pairs provide 320 training and 80 validation pairs for reasoning-trace analysis, with validation selecting the analysis layer. We retain a pair only if both traces have a parsed answer and usable checkpoints. Sample counts can therefore differ by model. Table~\ref{tab:reasoning_checkpoints} defines the five checkpoints.

\begin{table}[!htbp]
\centering\small
\renewcommand{\arraystretch}{1.05}
\setlength{\tabcolsep}{4pt}
\caption{\textbf{Reasoning-trajectory checkpoints.} $T$ counts reasoning tokens, and positions are one-based.}
\label{tab:reasoning_checkpoints}
\begin{tabular*}{\textwidth}{@{\extracolsep{\fill}}lp{0.77\textwidth}@{}}
\toprule
Checkpoint & State position \\
\midrule
\textsc{Pre} & Last prompt token before reasoning \\
\textsc{Early} & Reasoning token $\lfloor T/3\rfloor$ \\
\textsc{Middle} & Reasoning token $\lfloor 2T/3\rfloor$ \\
\textsc{Late} & Last reasoning token, $T$ \\
\textsc{Answer} & Token immediately preceding the first final-choice token \\
\bottomrule
\end{tabular*}
\end{table}

Validation selects one layer per model: L17 for Qwen3.5-4B, L16 for Qwen3.5-9B, L1 for Gemma-4-E4B, and L20 for Gemma-4-12B. Test samples do not affect this choice. The selected depth differs across models, so one layer cannot represent the Composite answer equally well in every architecture.

MMStar and RealWorldQA use separate early-stopping splits, described in Appendix~\ref{app:early_stopping}.

\section{Early-Stopping Details}
\label{app:early_stopping}

\subsection{Datasets and Cross-Validation}

We use every MMStar and RealWorldQA question in two-fold cross-validation. Within one fold, we train the detector and select its threshold; we test on the other fold (Table~\ref{tab:stopping_splits}). All checkpoints from one question stay together, as do RealWorldQA questions that share an image. Every model uses the same folds.

\begin{table}[!htbp]
\centering
\small
\renewcommand{\arraystretch}{1.05}
\caption{\textbf{Early-stopping evaluation splits.} Counts are questions. Each source fold is divided into detector training and threshold validation, and the other fold is held out for testing.}
\label{tab:stopping_splits}
\begin{tabular*}{\textwidth}{@{\extracolsep{\fill}}llrrr@{}}
\toprule
& & \multicolumn{2}{c}{Source fold} & Held-out fold \\
\cmidrule(lr){3-4}\cmidrule(l){5-5}
Dataset & Evaluation & Training & Validation & Test \\
\midrule
\multirow{2}{*}{MMStar}
& A$\to$B & 600 & 150 & 750 \\
& B$\to$A & 600 & 150 & 750 \\
\addlinespace[3pt]
\multirow{2}{*}{RealWorldQA}
& A$\to$B & 306 & 77 & 382 \\
& B$\to$A & 305 & 77 & 383 \\
\bottomrule
\end{tabular*}
\end{table}

Each question contributes one held-out prediction per condition. Pooled accuracy weights questions equally across the two evaluation directions.

\section{Additional Early-Stopping Results}

The main text combines the two test folds and analyzes answer completion. Here we report each fold separately.

\subsection{Performance across Evaluation Folds}
\label{app:early_stopping_realworldqa}

Tables~\ref{tab:early_stopping_folds} and~\ref{tab:early_stopping_realworldqa_folds} give accuracy and mean reasoning length for each test fold. Reasoning length falls in every model and dataset, but accuracy changes differ between folds.

\begin{table}[!htbp]
\centering
\small
\renewcommand{\arraystretch}{1.05}
\setlength{\tabcolsep}{2pt}
\caption{\textbf{Per-direction held-out MMStar results.}
Each row evaluates 750 held-out cases.
Accuracy values are percentages and reasoning-token counts are means.
$\Delta$ denotes the change from full thinking to adaptive stopping in percentage points.}
\label{tab:early_stopping_folds}

\begin{tabular*}{\textwidth}{
    @{\extracolsep{\fill}}
    l
    c
    ccc
    cc@{}
}
    \toprule
    &
    & \multicolumn{3}{c}{\textbf{Accuracy (\%)}}
    & \multicolumn{2}{c}{\textbf{Reasoning Tokens}} \\
    \cmidrule(lr){3-5}
    \cmidrule(lr){6-7}
    Model
    & Train$\to$Test
    & Full
    & \textbf{Stop}
    & $\Delta$
    & Full
    & \textbf{Stop} \\
    \midrule

    \multirow{2}{*}{Qwen3.5-4B}
    & A$\to$B
    & 64.13
    & 64.13
    & $0.00$
    & 2,101
    & \textbf{256} \\
    & B$\to$A
    & 65.60
    & \textbf{65.73}
    & $\uparrow\,0.13$
    & 1,982
    & \textbf{254} \\
    \addlinespace[2pt]

    \multirow{2}{*}{Qwen3.5-9B}
    & A$\to$B
    & 66.93
    & \textbf{69.73}
    & $\uparrow\,2.80$
    & 1,979
    & \textbf{256} \\
    & B$\to$A
    & 66.80
    & \textbf{70.00}
    & $\uparrow\,3.20$
    & 1,922
    & \textbf{254} \\
    \addlinespace[2pt]

    \multirow{2}{*}{Gemma-4-E4B}
    & A$\to$B
    & 57.47
    & 57.47
    & $0.00$
    & 767
    & \textbf{437} \\
    & B$\to$A
    & 58.67
    & \textbf{59.87}
    & $\uparrow\,1.20$
    & 739
    & \textbf{733} \\
    \addlinespace[2pt]

    \multirow{2}{*}{Gemma-4-12B}
    & A$\to$B
    & 57.60
    & \textbf{66.67}
    & $\uparrow\,9.07$
    & 1,822
    & \textbf{327} \\
    & B$\to$A
    & 56.27
    & \textbf{64.93}
    & $\uparrow\,8.67$
    & 1,885
    & \textbf{246} \\

    \bottomrule
\end{tabular*}
\end{table}

\begin{table}[!htbp]
\centering
\small
\renewcommand{\arraystretch}{1.05}
\setlength{\tabcolsep}{2pt}
\caption{\textbf{Per-direction held-out RealWorldQA results.}
A$\to$B evaluates 382 questions, and B$\to$A evaluates 383.
Accuracy values are percentages and reasoning-token counts are means.
$\Delta$ denotes the change from full thinking to adaptive stopping in percentage points.}
\label{tab:early_stopping_realworldqa_folds}

\begin{tabular*}{\textwidth}{
    @{\extracolsep{\fill}}
    l
    c
    ccc
    cc@{}
}
    \toprule
    &
    & \multicolumn{3}{c}{\textbf{Accuracy (\%)}}
    & \multicolumn{2}{c}{\textbf{Reasoning Tokens}} \\
    \cmidrule(lr){3-5}
    \cmidrule(lr){6-7}
    Model
    & Train$\to$Test
    & Full
    & \textbf{Stop}
    & $\Delta$
    & Full
    & \textbf{Stop} \\
    \midrule

    \multirow{2}{*}{Qwen3.5-4B}
    & A$\to$B
    & 71.73
    & \textbf{74.87}
    & $\uparrow\,3.14$
    & 1,692
    & \textbf{247} \\
    
    & B$\to$A
    & 73.89
    & \textbf{75.98}
    & $\uparrow\,2.09$
    & 1,700
    & \textbf{322} \\
    \addlinespace[2pt]
    \multirow{2}{*}{Qwen3.5-9B}
    & A$\to$B
    & 73.30
    & \textbf{78.27}
    & $\uparrow\,4.97$
    & 1,521
    & \textbf{252} \\
    
    & B$\to$A
    & 72.58
    & \textbf{77.55}
    & $\uparrow\,4.96$
    & 1,692
    & \textbf{244} \\
    \addlinespace[2pt]
    \multirow{2}{*}{Gemma-4-E4B}
    & A$\to$B
    & \textbf{57.33}
    & 56.81
    & $\downarrow\,0.52$
    & 448
    & \textbf{347} \\
    
    & B$\to$A
    & \textbf{59.01}
    & 57.44
    & $\downarrow\,1.57$
    & 436
    & \textbf{344} \\
    \addlinespace[2pt]
    \multirow{2}{*}{Gemma-4-12B}
    & A$\to$B
    & 59.95
    & \textbf{63.87}
    & $\uparrow\,3.93$
    & 781
    & \textbf{217} \\
    
    & B$\to$A
    & 60.31
    & \textbf{69.71}
    & $\uparrow\,9.40$
    & 936
    & \textbf{369} \\

    \bottomrule
\end{tabular*}
\end{table}

\end{document}